# Rapid Assessments of Light-Duty Gasoline Vehicle Emissions Using On-Road Remote Sensing and Machine Learning

Yan Xia[1], Linhui Jiang[1], Lu Wang[1], Xue Chen[1], Jianjie Ye[5], Tangyan Hou[1], Liqiang Wang[1], Yibo Zhang[1], Mengying Li[1], Zhen Li[1], Zhe Song[1], Yaping Jiang[1], Weiping Liu[1], Pengfei Li[3+], Daniel Rosenfeld[4], John H. Seinfeld[2], Shaocai Yu[1,2+]

[1]Research Center for Air Pollution and Health; Key Laboratory of Environmental Remediation and Ecological Health, Ministry of Education, College of Environment and Resource Sciences, Zhejiang University, Hangzhou, Zhejiang 310058, P.R. China
[2]Division of Chemistry and Chemical Engineering, California Institute of Technology, Pasadena, CA 91125, USA.
[3]College of Science and Technology, Hebei Agricultural University, Baoding, Hebei 071000, P.R. China
[4]Institute of Earth Sciences, The Hebrew University of Jerusalem, Jerusalem, Israel
[5]Bytedance Inc., Hangzhou, Zhejiang 310058, China

[+]*Correspondence to*: Shaocai Yu (shaocaiyu@zju.edu.cn); Pengfei Li (lpf_zju@163.com)

**Synopsis.**

On-road remote sensing measurements, assisted by machine-learning, can realize day-to-day supervision of urban vehicle emissions.

**TOC.**

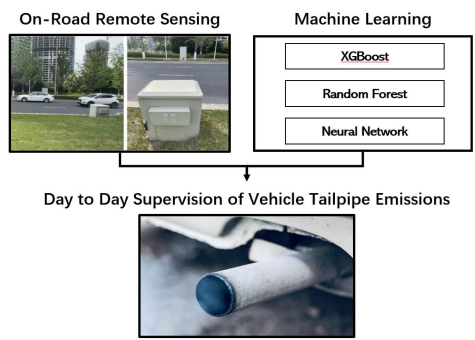

**Abstract.**

In-time and accurate assessments of on-road vehicle emissions play a central role in urban air quality and health policymaking. However, official insight is hampered by the Inspection/Maintenance (I/M) procedure conducted in the laboratory annually. It not only has a large gap to real-world situations (e.g., meteorological conditions) but also is incapable of regular supervision. Here we build a unique dataset including 103831 light-duty gasoline vehicles, in which on-road remote sensing (ORRS) measurements are linked to the I/M records based on the vehicle identification numbers and license plates. On this basis, we develop an ensemble model framework that integrates three machining learning algorithms, including neural network (NN), extreme gradient boosting (XGBoost), and random forest (RF). We demonstrate that this ensemble model could rapidly assess the vehicle-specific emissions (i.e., CO, HC, and NO). In particular, the model performs quite well for the passing vehicles under normal conditions (i.e., lower VSP (< 18 kw/t), temperature (6 ~ 32 °C), relative humidity (< 80%), and wind speed (< 5m/s)). Together with the current emission standard, we identify a large number of the 'dirty' (2.33%) or 'clean' (74.92%) vehicles in the real world. Our results show that the ORRS measurements, assisted by the machine-learning-based ensemble model developed here, can realize day-to-day supervision of on-road vehicle-specific emissions. This approach framework provides a valuable opportunity to reform the I/M procedures globally and mitigate urban air pollution deeply.

# 1 Introduction

Global urbanization and vehicle population grow mutually[1,2]. A common consequence is the excessive dependence on fossil fuels and thus urban air pollution[3,4]. Particularly, on-road vehicle emissions, like carbon monoxide (CO), nitric oxide (NO), and Hydrocarbons (HCs), play a key role in forming atmospheric ozone and fine particulate matter, resulting in global health concerns[5–8]. Hence, in-time and accurate assessments of on-road vehicle emissions are essential for air quality and epidemiological policymaking.

However, official insight primarily relies on required Inspection/Maintenance (I/M) procedures for vehicle-specific emissions, which are incapable of regular (e.g., daily) real-world supervision of a large gap to the real-world conditions (e.g., meteorological conditions)[9,10]. Early measurement studies have presented a consensus that there is a heavy-tail distribution of on-road vehicle emissions. The top 10% of vehicles are typically responsible (i.e., > 50%) for the total emissions of the vehicle fleet, while the overwhelming majority emit minimal emissions, regardless of model years[10,11]. With this background, there is thus a valuable opportunity for air quality managements to screen real-world vehicles regularly. On this basis, to determine whether to oblige them to retake a part in the I/M procedures or exempt them is a potential achievement of policy reform.

To this end, numerous instrument approaches have been increasingly applied to assess on-road vehicle emissions at a high resolution globally, such as On-Board Diagnosis (OBD)[12], On-Board Portable Emissions Measurement (OBPEM)[13], and On-Road Remote Sensing (ORRS)[14]. The OBD technique is a comprehensive electronic system embedded in vehicles, collecting real-world operating data but unable to report emissions. By comparison, the PEM method is also embedded but directly measures the vehicle emissions. Yet, they both have not been universally applied due to huge expenses. To date, ORRS has been routinely applied in the United States for decades[15], followed by Europe (e.g., United Kingdom, Spain, Sweden, and Switzerland)[16–20] and, more recently, in China[21–24]. ORRS can provide economic and promising measurements for on-road vehicle emissions. Typical ORRS campaigns last from days to weeks, with up to millions of records, and recent advances combine different ORRS campaigns to explore long-term ORRS records (e.g., for decades).

However, there is a knowledge gap in reliably inspecting the emissions for a certain vehicle (i.e., vehicle-specific emissions) using ORRS. Multiple real-world drivers affect the ORRS consequences nonlinearly and substantially, primarily including the meteorological conditions (e.g., relative humidity and pressure)[10,14,25–28]. As a result, previous studies generally found a large gap in the measurements between ORRS and PEM or the I/M procedures, such that neither PEMs or the I/M procedures can directly validate ORRS results. In turn, relying solely on ORRS, without the consideration of real-world drivers, is prone to misidentify high- or low-emitting vehicles.

To address this issue, scientists make considerable efforts to develop modelling approaches. These models fall into several classes, including simple or compound regression models[29], chemical transport models[27], and, most recently, machine learning models[30], each with different strengths and limitations. As an emerging technique, machine learning, such as neural network (NN)[31], random forest (RF)[32], and extreme gradient boosting (XGBoost)[33], has a great advantage in representing nonlinear relationships, conducive to enhancing the prediction ability in terms of accuracy. Hence, given the complex impacts of those real-world drivers on ORRS, machine learning algorithms offer a potential alternative to characterize the nonlinear relationships between the ORRS and I/M measurements. On this basis, ensemble approaches have been developed to integrate diverse base machine learning algorithms and incorporate their predictions[34–36]. In so doing, it is projected that they could exploit multiple predictive power and thus outperform the performance of individual algorithms. Wolpert et al. first demonstrated the combination of multiple statistical models that further enhanced the robustness and accuracy of the base models[37]. This kind of method emerged in spatiotemporal issues but has not been extended to understand the ORRS measurements by far, or to grade vehicles in terms of emissions. Recent studies generally focus on air pollution prediction, which apply machine-learning-based ensemble models to predict the spatiotemporal distributions of air pollutant concentrations over different regions (e.g., China, Europe, and the United States), and, as expected, exhibit an advantage over the individual models and high performance[34–36].

Here, we address a unique dataset that covers a large count of the ORRS-based, I/M-based vehicle-specific emission measurements and the associated real-world drivers (e.g., the meteorological conditions). The information is in one-to-one correspondence and comprehensively covers a broad spectrum of light-duty gasoline vehicles in China and a wide range of real-world states. On this basis, we develop an ensemble model framework that integrates three machining learning algorithms (i.e., NN, RF, and XGBoost). The objective is to rapidly assess on-road vehicle-specific emissions and further distinguish dirty or clean vehicles. The underlying assumption is that each algorithm has its exclusive merits and limitations[34]. The consequent ensemble model thus incorporates these heterogeneous predictions and is beneficial for our objective. We evaluate our model performance by exploring its ability to assess on-road vehicle-specific emissions. Our results demonstrate that the ORRS, assisted by the machine-learning-based ensemble model, can realize day-to-day supervisions of on-road vehicle emissions and thus be central to urban air pollution mitigation.

## 2 Materials and Methods

### 2.1 On-Road Remote Sensing Measurements

We conducted a long-term ORRS field campaign (i.e., from May 2020 to November 2020) at two roadside sites in Hangzhou (Fig. S1). We made data collection (i.e., 114356 records) for 103831 light-duty gasoline vehicles (i.e., LGVs). Note that these vehicles were the results of matching with the I/M dataset, as detailed in Sect. 2.2. Each record contained a set of the real-world activities and emission information. Specifically, for LGV, CO, HC, and NO emissions, instantaneous speeds and accelerations were measured concurrently. This ORRS technique has been used and discussed extensively in previous studies. First, a remote vehicle tailpipe exhaust sensor, as the key core of ORRS, consists of a light detector situated on one side of the road and a receiver on the other side. The former was equipped with two dispersive ultraviolet spectrometers for NO (198 ~ 227 nm) and four nondispersive selenide infrared sensors for CO (3.6 μm), $CO_2$ (4.3 μm), HC (3.3 μm), and a reference channel (3.9 μm). In theory, the light transmitted through the vehicle exhaust plumes would be attenuated proportionally with the gas-phase emission concentrations. Consequently, each gas was detected as a molar ratio to $CO_2$ (i.e., $CO/CO_2$, $HC/CO_2$, $NO/CO_2$). We also applied an external reference gas to conduct regular calibrations in order to avoid possible instrument drifts caused by atmospheric pressure, temperature, or background pollutants. This reference gas was stored in three calibration cylinders, containing 6% CO, 0.3% NO, 0.6% propane, and 15% $CO_2$ and certified to ± 2% accuracy.

For the passing vehicles, instantaneous speeds and accelerations were also measured via a pair of parallel infrared beams. The detectors were approximately 0.3 meters above the ground and 10 meters apart. Simultaneously, their license plates were extracted automatically via an image recognition algorithm based on their video images. Once the local vehicles passed the sample sites, their detailed information could be retrieved from the official I/M database, which we discussed in the following section. Along with the ORRS-based emissions and operating modes, meteorological factors, including temperature, relative humidity, wind speed, and pressure, were recorded. Note that, once relative humidity and temperature exceed 95% and 40°C, respectively, the ORRS measurements would lack robustness and thus be eliminated. The occurrence probability of this case was 6.56% (i.e, 7498 records).

**2.2 Vehicle Inspection/Maintenance Measurements**

The official dataset of the I/M measurements provided a collection of records for 153481 LGVs registered in Hangzhou. On request, the vehicle-specific static attributions need to be inspected annually, including vehicle identification number (VIN), model year, accumulated mileage, capacity, wheelbase, and maximum horsepower. A large number of studies have proved that these underlying features are closely related to the final emission magnitudes and should thus be considered in the emission models. Along with these static features, the vehicle-specific emissions, including HC, CO, and NO, were measured via the probe in the tailpipe. For a certain vehicle, its static features and emission measurements were both assembled into a single record based on the unique VIN and license plate.

**2.3 Model Building and Evaluation**

We developed an ensemble model that combined multiple machine-learning models to optimize the estimates of real-world vehicle-specific emissions via the ORRS measurements. Such a model exploited the advantages of each model and potentially outperformed the individual. The following two stages have been used to develop the ensemble model: (i) We predicted the real-world vehicle-specific emissions using three individual models: NN[31], RF[32], and XGBoost[33]; (ii) We then fitted an ensemble model based on the cross-validated estimations from those three models.

**2.3.1 Machine-Learning Models**

The ensemble model depended on three representative machine-learning methods, i.e., NN, RF, and XGBoost, to reconstruct an ensemble of predictions from different machine-learners. First, we adopted a fully connected NN learner (FCNN)[38] and established two hidden layers, both with 256 nodes. In theory, the FCNN has a powerfully learning ability but possibly overfits, leading to a remarkable variance in predictions. To prevent overfitting, we applied a dropout layer for each dense layer, which is a technique to randomly drop units from the fully connected neural network during training[39]. Second, RF offered distinct advantages by comparison because the sampling process made full use of the randomness[32]. Therein each decision tree was built upon a bootstrap sample of the data and enabled the successive input variables. When the desirable number of trees were trained completely, the average of their outputs would be treated as the final ensemble prediction. Given this mechanism, an increasing number of trees would inevitably lead to more time-consuming training.

Third, XGBoost was developed based on the gradient boosting decision trees[33]. In contrast to RF, where the trees were independent, such trees in XGBoost were dependent. In theory, each new tree was built to predict the residuals of prior trees. This indicates that it focused on investigating the performance of the previous trees. Then the total trees were collected to derive the final predictions. By comparison, XGBoost was of more efficient training processes and thus widely used in data mining competitions.

We applied Grid Search to optimize the hyperparameters of FCNN, RF, and XGBoost, which was a common technique to obtain optimal structures as listed in Table S1[40]. For FCNN, the optional learning rates were $10^{-3}$, $10^{-2}$, and $10^{-1}$, while the alternative dropout rates were 0.1, 0.2 and 0.3. The epochs ranged from 100 to 400. We selected ReLU as the activation functions in the hidden layers and the output layer, respectively, while the loss function was unified as the root-mean-square error (RMSE). For RF, the search space was: the maximum tree depths from 3 to 8, the criterion an either-or option (i.e., Gini or Entropy), the minimum sample split from 2 to 6, and the number of trees among 100, 200, 300, and 400[41]; For XGBoost, the search space was: the maximum tree depths from 4 to 8; the minimum child weights from 1 to 5; the subsample ratios from 0.5 to 0.9; and the subsample ratio of columns from 0.5 to 0.9. On this basis, we needed to avoid these machine learning models being overfitted. Hence, only the parameters that were of high evaluation statistic Gains (> 0.01) were considered. This index indicates the degree of the accuracy improvement after splitting on the corresponding feature.

**2.3.2 Ensemble Model**

We assumed that any individual algorithm has its own advantages and limitations. To this end, we applied an ensemble method based on stacked generalization to enhance the strengths of different machine learning algorithms and to offset their weaknesses. The objective is to further optimize the predictions of the vehicle-specific emissions via the individual machine learners (FNN, RF, and XGBoost).

Here, a two-layer stacked ensemble model was proposed (Fig. S2). The first layer assembled FCNN, RF, and XGBoost methods to exploit original vehicle-specific features and achieve initial predictions. It can be viewed as a means of integrating the errors of all base generalizers when working on a particular learning set. To correct the prediction residuals and avoid overfitting, we further trained a higher but simpler meta-model (i.e., XGBoost), generating final predictions for each pollutant from a certain vehicle. Hence, the over framework of this ensemble model could be regarded as a super multi-layer perceptron that utilized the individual model as neural units of the hidden layer and the meta-model as the final units of the output layer with the objective of maximizing prediction accuracy, generalizability, and robustness.

### 2.3.3 Mechanistic Interpretation

For complex machine-learning models, mechanistic interpretations regarding how features affect predictions is often desirable and at times mandatory. Here, to understand the mechanistic principles of the ensemble model, we calculated the Shapley values to quantify the contributions of the input factors[42,43]. The detailed information was illustrated in SI. As a result, a negative or positive Shapley value signified that the corresponding feature pushed the predictions towards the negative or positive directions, respectively. In this study, there were 25 input factors that were considered in the above machine learning models, as summarized in Table S2, thus corresponding to 25 Shapley values. Hence, the mean Shapley (MS) value for the feature *i* represented the positive and negative contributions of the input factor (i.e., independent variable) to the predicted vehicle-specific results (i.e., dependent variables). Besides, all Shapley values for a feature *i* were used to calculate its mean absolute Shapley (MAS) value. The overall impacts of each input factor can thus be quantified by the MAS values. The higher the MAS value, the more substantial the factor that affects the model estimations.

### 2.3.4 Model Evaluation

The ORRS measurements of on-road vehicle emissions served as input variables of the three machine learning algorithms. The matched I/M measurements based on the VINs were the dependent variables. They were split into the training and testing groups and thus cross-validated the model performance. Given that we first performed the individual machine-learners and then implemented an ensemble model, the cross-validation was designed separately to avoid overfitting. The model validation was operated through two substages (Fig. S2):

(1) We applied the 10-fold cross-validation to evaluate the performance of the individual machine learning models (i.e., FNN, RF, and XGBoost). Specifically, we divided all the I/M measurements into ten splits. We first applied the individual machine learning model (i.e., FNN, RF, and XGBoost) to predict the vehicle-specific emissions for 10% of the I/M measurements. Besides, 80% of the I/M measurements were used to train these models. The remaining 10% was kept aside for cross-validating the results of the individual machine learning models.

(2) In the second sub-stage, the cross-validated results and predictions in the first substage were integrated into the new training and test datasets, respectively. On this basis, the predictions of the meta leaner (i.e., XGBoost) would serve as the final predictions of the ensemble model.

As a result, all the predictions can be compared to the I/M measurements. To understand the model performance, we utilized four statistic indexes, including coefficient of determination ($R^2$), RMSE, Slope, and Intercept, to evaluate the predictions. The significances of these indexes were described as follows[35]:

(i) $R^2$: The measured and predicted vehicle-specific emissions, including HC, CO, and NO emissions, were regressed. On this basis, the explained variances were calculated;

(ii) RMSE: This parameter represents a summary measure of the prediction errors. It was calculated as the square root of the mean quadratic differences between the measured and predicted vehicle-specific emissions;

(iii) Slope and Intercept: The two parameters represent the multiplicative bias and the additive bias of the predictions. They were calculated as the coefficient and intercept from the linear regression between the measured and predicted vehicle-specific emissions.

**2.3.5 Model Robustness**

On the basis of the reliable predictions, we could empirically find two thresholds that can screen the over standard and standards-compliant vehicles. The standards came from the official guideline[44,45], e.g., 8 g/km for CO, 1.6 g/km for HC, and 1.3 g/km for NO. Therefore, a more flexible policy for vehicle emission controls can be derived. At first glance, the standards-compliant vehicles might not need to participate in the annual I/M procedures (i.e., Free-I/M vehicles). In contrast, the over standard ones would be required to be re-inspected as soon as possible (i.e., Re-I/M vehicles). Note that, for a certain vehicle, it would be treated as a Re-I/M vehicle once any pollutants were over the standards, while only if all pollutants were up to the standards could it be recognized as a Free-I/M vehicle. The rest of the vehicles need to go through the I/M procedures regularly.

Hence, we need to investigate the robustness of the ensemble model in classifying the vehicles. We employed a Monte Carlo approach[46] that included **T**-fold random sampling simulations. Here, given sufficient **T** repetitions, e.g., 500 times, it was possible to investigate the statistical robustness of those thresholds. Specifically, in each sampling simulation, we randomly selected different sub-datasets as the alternative cross-validation dataset. Each sub-dataset had **n** records that depended on the actual demands, the ratios of the over standard vehicles to the standards-compliant ones (i.e., in term of the official guideline) were equal to the full dataset. Previous studies have demonstrated that such bootstrap samplings could be used to asymptotically infer the true thresholds and the robustness of our ensemble model[46].

To quantify the robustness of the predicted Free-I/M or Re-I/M thresholds for a given pollutant, we applied two statistic indexes, including the relative error (RE) and absolute error (AE), which were calculated as follows:

$$RE = |\bar{x_i} - x| / x \times 100\% \quad (Eq.2),$$

$$AE = |\bar{x_i} - x| \quad (Eq.3).$$

Therein $x$ denoted the predicted threshold based on the universal dataset, which was treated as the reference threshold. The **T**-time simulations generated a series of the predicted thresholds $x_1, \ldots, x_i, \ldots, x_n$ based on the sample sub-datasets. The $\bar{x_i}$ is the average predicted thresholds of the **T**-time simulations. Lower **RE** and **AE** corresponded to greater confidence of the predictions.

## 3 Results and Discussion

### 3.1 Model Validation

Figure S3 shows the ORRS-based and I/M-based vehicle-specific relationships between the emissions (including CO, HC, and NO) and the vehicle-specific features (e.g., Accumulated Mileages and Model Years). As shown in the descriptive statistics, the median of the vehicle-specific HC emissions from the former was 15 ppm, while that from the latter was 0.05 g/km. Despite this, it is immediately clear that the emissions derived from different measurements (i.e., the ORRS and I/M measurements) presented similar distributions across those features. For instance, from the perspective of accumulated mileages, most (80%) of the ORRS-based vehicle-specific HC emissions ranged from 0 ~ 63 ppm, while most of the I/M-based results ranged from 0.01 ~ 0.12 g/km. This indicates that the machine learning models have a large potential to mine the quantitative relationships between the ORRS-based and I/M-based datasets. In addition, such relationships were also affected substantially by the meteorological conditions. These factors, like temperature, relative humidity, wind speed, and pressure, were summarized in Table S2. For instance, the ORRS-based vehicle-specific emissions responded strongly to the varying temperature, relative humidity, and wind speed (Fig. S4). Thus, it is projected that all these factors likewise played a key role and should serve as the key input data of the individual machine-learning models.

Figure 1 presents the cross-validation of the predicted vehicle-specific emissions (including CO, HC, and NO) via NN, RF, XGBoost, and their ensemble model. The detailed statistics were also summarized. We found that the performance of the single machine learning models varied across pollutants (i.e., HC, CO, and NO) but was quite close for a certain pollutant. For HC and CO, the cross-validation $R^2$ of XGBoost (0.86, 0.84) was significantly higher than that of FCNN (0.81, 0.82) or RF (0.76, 0.80). Also, XGBoost was of the lowest RMSE (0.12, 1.97), and the closest to 0 Intercept (0.00, -0.10), and the closest to 1 Slope (1.04, 1.04), indicating a minimum bias among the individual models. In contrast, the NN model performed the best (RMSE: 0.20; $R^2$: 0.85; Intercept: 0.04; Slope: 0.96) for NO, followed by XGBoost (RMSE: 0.22; $R^2$: 0.81; Intercept: 0.05; Slope: 1.03) and XX (RMSE: 0.26; $R^2$: 0.74; Intercept: 0.10; Slope: 1.06). More importantly, for any vehicle-specific emission pollutants, the overall model performance of the ensemble model outperformed those from any single algorithm, with the lowest RMSE (0.11, 0.18, 1.90), the highest $R^2$ (0.93, 0.88, 0.89), the closest to 0 Intercept (0.01, 0.00, -0.03), the closest to 1 Slope (1.01, 1.01, 1.02) (for HC, NO, and CO, respectively). Collectively, our findings support the initial hypothesis that an ensemble model might have an advantage over the individual models. This is due to its ability to incorporate multiple predictions from different machine-learners.

To further explore the model performance, we investigated the responses of the ensemble model performance to the varying vehicle-specific features. Figure S5 shows the cross-validated NO results of different models across different vehicle-specific features, including wheel base, torsion, fuel tank capacity, and maximum horsepower. For each vehicle-specific feature, the vehicle population consistently presented an approximately normal distribution. On this basis, three bins were divided according to the magnitudes of the feature. Overall, we found that the performance of the models varied across different features. This exhibition was expected given their potentially distinct roles in affecting the model estimations. It should be noted that, for each case, the ensemble model consistently had the best performance. On the other hand, even for each feature, the performances of the models varied across different bins because of different input amounts. Hence, the peak bin contained the most input data, while the bottom bin had the least data, as shown in Fig. S5. By comparison, the ensemble model for the peak bin performed the best, particularly outperforming the three machine learning models for other bins. For instance, in terms of the wheel base, the predictions of the ensemble model for the peak bin had the highest $R^2$ (0.91) and the lowest RMSE (0.17), and those for the bottom bin also presented the similar performance, with the highest $R^2$ (0.87) and the lowest RMSE (0.19). Performance was relatively weak, but still excellent, for the peak bin of the torsion (RMSE: 0.18; $R^2$: 0.88; Intercept: 0.00; Slope: 1.00) and the bottom bin of the maximum horsepower (RMSE: 0.17; $R^2$: 0.88; Intercept: 0.01; Slope: 1.01) vehicles**.** In terms of wheel base, the ensemble model had the best performance for vehicles with high wheel base with $R^2$ equal to 0.91. In terms of maximum horsepower, the ensemble model results show relatively little variations with $R^2$ ranging from 0.87 to 0.88, RMSE ranging from 0.17 to 0.19. Similar performances were also found in the cross-validated CO and HC results (Figs. S6 and S7). Therefore, the ensemble model classified by any vehicle-specific attributions outperformed the individual machine learning models.

Finally, the remaining test dataset was used to evaluate the predictive power of the ensemble model (Fig. S8). Overall, the predicted vehicle-specific emissions were close to the I/M results. For instance, the predicted results for HC and NO were of robust evaluation statistics (RMSE: 0.12; $R^2$: 0.94; Intercept: 0.02; Slope: 0.95 for HC, and RMSE: 0.21; $R^2$: 0.89; Intercept: -0.01; Slope: 1.03 for NO). By comparison, the ensemble model performed better on CO with a steeper slope (1.02), increased $R^2$ (0.88), decreased RMSE (2.01), and decreased Intercept (0.02). To sum up, our hundreds of randomized trials on the model trainings, CVs, and evaluations show that the prediction results of the ensemble model can be potentially treated as rapid inspections for vehicle-specific emissions.

**3.2 Mechanistic Interpretation**

Figure 2 illustrates the MS and MAS values for each input variable for the predicted vehicle-specific emissions (i.e., CO, HC, and NO) via the ensemble model. For a certain feature, its contributions to the prediction were not changeless, or even overturned (Fig. S9). At some time, the temperature, relative humidity, and wind speed made positive contributions to the final predictions, as defined by the positive Shapeley values. At other times, the meteorological input variables made negative contributions to the results, as indicated by the negative Shapley values. After summing these positive and negative Shapley values, we obtained the net MS values (Fig. 2). The positive net MS values occurred in acceleration, velocity, fuel tank capacity, vehicle volume, temperature, and model year. This indicates that these input variables contributed positively to the ensemble model. In particular, acceleration made the most significant positive contribution to the predicted vehicle-specific emissions, followed by velocity, fuel tank capacity, vehicle volume, temperature and model year. In contrast, accumulated mileage, VSP, RS_NO, and wind speed made the largest negative contributions. Besides, the contributions of maximum horsepower, relative humidity, torsion, and engine type were either positive in some cases or negative in others, and their summed net MS values were small. It should be noted that such positive and negative patterns of these MS values kept stable for different pollutants.

From another perspective, for any pollutant, RS_HC, RS_CO, RS_NO, accumulated mileage, VSP, acceleration, and velocity were consistently the most influential factors that exhibited the largest MAS values (Fig. 2). This result was consistent with the traditional mechanisms in a large number of bottom-up emission models. Nevertheless, there were substantial differences in the MAS values between different pollutants. For CO, the MAS values of the RS_CO, RS_HC, RS_NO, Velocity, VSP, and accumulated mileage were up to 2.70, 2.63, 1.97, 1.09, 0.67 and 0.52, respectively, significantly higher than those (< 0.5) of other factors (e.g., total mass, relative humidity, and wind speed). For NO and HC, RS_HC, RS_CO, RS_NO, and accumulated mileage yielded much more contributions (> 0.08 and >0.11, respectively) to the predicted results. This indicates that RS_HC, RS_CO, RS_NO, accumulated mileage, and VSP might still dominate the ORRS-based vehicle-specific emissions. Furthermore, meteorological factors cannot be neglected in the ensemble model, especially for NO and HC.

**3.3 Inspection of vehicle-specific emissions**

As expected, the ensemble model did not perform equally well at different emission magnitude levels, especially for high and low levels (Fig. S8). Here we first distinguished the Free-I/M and Re-I/M vehicles from the perspective of each pollutant (i.e., CO, HC, and NO). Figure 3 shows the over-standard rates of the predicted results across the I/M-based emission magnitudes. For NO, the over-standard rate was up to 100% or touched the bottom (0%) if the predicted magnitudes were larger than 0.59 g/km or lower than 2.88 g/km, much stricter than current official standards. As a result, three intervals divided by the two thresholds were achieved. The former interval corresponded to the Free-I/M vehicles, covering 5839 vehicles (60.61%), while the latter one corresponded to the Re-I/M vehicles (0.70%). For CO and HC, the intervals of the Free-I/M and Re-I/M vehicles varied to some extent. For instance, the Free-I/M vehicles decreased to 56.90% or increased to 66.66%. Collectively, a wide range of vehicles fell into the intervals of the Free-I/M (48.27%) vehicles. By comparison, the identified Re-I/M vehicles were much less (1.01%), but still covered a large number (i.e., dozens of vehicles).

The exact threshold likely depended on the cross-validation dataset, i.e., the testing group of the I/M measurements, with different settings probably leading to distinct thresholds. Using the Monte Carlo approach, we randomly selected different sub-dataset of the I/M measurements as the alternative cross-validation dataset. Each sub-dataset had records of 15000, 15.57% of the total dataset. Figure S10 presents that the thresholds for both Free-I/M and Re-I/M vehicles fluctuated narrowly (< 6.35%). Therefore, the predicted thresholds were robust on the whole and would not be influenced significantly by the specific I/M measurements.

Besides, the exact threshold was also closely related to the ORRS measurements that were affected by the ORRS sample sizes, vehicle-specific features, and meteorological conditions substantially. First, the **RE**s and **AE**s of the predicted NO thresholds showed as a function of the ORRS sample sizes (Fig. 4). As expected, a larger sample size led to higher confidence in predicting the thresholds. More importantly, we found the tipping points in this function. Below the tipping point, further increases in the sample size led to remarkable decreases in the **RE**s and **AE**s. Specifically, a sample size of approximately 10000 records was needed for the predicted Free-I/M thresholds, while a significantly large size of 14000 records for the predicted Re-I/M thresholds. Despite the stark difference in the sample size, a possible consequence of the distinct shape of the Free-I/M and Re-I/M vehicle distributions, the sample size was still much less than the ORRS dataset size. Moreover, this rule was likely suitable for any vehicle-specific emissions (i.e., CO and HC) (Figs. S11 and S12). Second, vehicle-specific velocity, acceleration, vehicle mass, grad, and altitude were systematically considered in the definition of VSP[28]. This integrated parameter has been extensively investigated in the ORRS and model analyses. Figure S13 presents the variations in the predicted vehicle-specific thresholds under various ORRS-based VSP bins. We found that the responses of the predicted thresholds to the VSP variations were mostly similar and stable. Third, stable meteorological conditions with low temperature, relative humidity, and wind speed were conducive to our ensemble model. The standard deviations and tolerated errors of the predicted thresholds kept the high certainties and stable when the ambient temperature, relative humidity, and wind speed ranged from 6°C to 32°C, from 0% to 80%, and from 0 m/s to 5 m/s, respectively (Fig. S14).

Therefore, to further improve the ensemble model performance, we could restrict the predictions to those with sufficient sample sizes and filtered vehicle features and meteorological conditions. Overall, the ensemble model performed relatively well for the passing vehicles with low temperature, relative humidity, and wind speed. As illustrated above, we filtered out the predictions for the relative humidity below 80%, the wind speed below 5m/s, the temperature ranging from 6°C to 32°C (Fig. S15). As a result, a total of 83286 vehicles met the standard, accounting for 76.01% of the total. In this way, once the predicted vehicle-specific HC, NO, or CO emissions were larger than 2.18g/km, 2.32g/km, or 12.30g/km, respectively, the associated vehicles can be treated as the Re-I/M vehicles. At the opposite side, the Free-I/M vehicles were the ones with the lower predicted vehicle-specific HC, NO, and CO emissions (i.e., 0.62g/km for HC, 0.65g/km for NO, and 2.68g/km for CO). Consequently, the proportions of the Free-I/M and Re-I/M vehicles to the filtered vehicles were up to 74.92% and 2.33%, respectively, increasing significantly compared to those for the vehicles that were not filtered.

## 4 Environmental Implications

In summary, we combine the ORRS measurements, a large-scale I/M database, and three machine learning algorithms to propose a reliable ensemble model. By interpreting the ensemble model based on the Shapley values, we further show that it is necessary to include RS_HC, RS_NO, RS_CO, VSP, accumulated mileage, acceleration, velocity, temperature, relative humidity, and wind speed to build reliable predictions. Relying on this reliable model, we can establish two thresholds to screen the Free-I/M and Re-I/M vehicles. For a given ensemble model framework, there is a practical need to strengthen the robustness of the predicted thresholds by increasing the ORRS sample size. On this basis, suitable VSP and meteorological conditions are also conducive. Note that these restrictive conditions are not exacting, presenting a good opportunity for the large-scale application of our approach. Therefore, our results are important for planning official ORRS campaigns and appropriate vehicle-specific I/M strategies.

If the objective of the ORRS campaigns is government surveillance of substandard vehicles, then more aggressive thresholds are needed. This is due to the relatively high model uncertainties in determining the Re-I/M threshold. Once a predicted vehicle-specific emission exceeds the substandard threshold, the corresponding vehicle is recognized as an illegal member and should be retested compulsively. In turn, to make our approach more inclusive, we might design the ORRS measurement campaign for exempting from the regular I/M procedures. More conservative strategies could be promoted owing to the relatively high model certainties. Whatever the target is, our approach will always capture very useful data for the whole vehicle fleet and even reform the regular I/M procedures. How robust and accurate the predicted results and thresholds are is essentially a measure of the statistical problem. To cooperate with more ORRS measurements might be helpful. Hence, either more ORRS units should be deployed to multiply the data capture, or they should be deployed to the road with higher traffic volume[47].

***Data availability.*** Three base machine-learning models are used in this study, including NN, RF, and XGBoost, which are obtained from the TensorFlow library, the Scikit Learn library, and the XGBoost library, respectively. Additional information and help are available by contacting the authors.

***Supplement.*** The supplement related to this article is available online.

Locations of two ORRS sample sites, model framework, the distributions of the ORRS-based and I/M-based measurements, the relationship between the ORRS-based vehicle-specific measurements and the varying meteorological conditions, the cross-validated results, MS and MAS values for input variables, variations in the thresholds for the Free-I/M and Re-I/M vehicle, ORRS-based VSPs vs the REs and AEs, Meteorological factors vs the REs and AEs, hyperparameters of the machine learning algorithms, input factors of the machine learning models and their Shapley (MS and MAS) values.

***Author contributions.*** S.Y., P.L., and Y.X. designed this research, developed the model, performed the analysis, and wrote the paper. L. J., L. W., X. C., T. H., L. W., Y. Z., M. L., Z. L., Z. S., Y. J., W. L., D. R., and J. H. S. made contributions to discussing and improving this research.

***Competing interests.*** The authors declare that they have no conflict of interest.

***Acknowledgements.*** This study is supported by the Department of Science and Technology of China (No. 2018YFC0213506 and 2018YFC0213503), National Research Program for Key Issues in Air Pollution Control in China (No. DQGG0107) and National Natural Science Foundation of China (No. 21577126 and 41561144004). Pengfei Li is supported by National Natural Science Foundation of China (No. 22006030), Initiation Fund for Introducing Talents of Hebei Agricultural University (412201904), and Hebei Youth Top Fund (BJ2020032)

**References**

(1) Kelly, F. J.; Zhu, T. Transport Solutions for Cleaner Air. *Science (80-. )*. **2016**, *352* (6288), 934 LP – 936. https://doi.org/10.1126/science.aaf3420.

(2) Gentner, D. R.; Jathar, S. H.; Gordon, T. D.; Bahreini, R.; Day, D. A.; El Haddad, I.; Hayes, P. L.; Pieber, S. M.; Platt, S. M.; de Gouw, J. Review of Urban Secondary Organic Aerosol Formation from Gasoline and Diesel Motor Vehicle Emissions. *Environ. Sci. Technol.* **2017**, *51* (3), 1074–1093.

(3) Grange, S. K.; Lewis, A. C.; Moller, S. J.; Carslaw, D. C. Lower Vehicular Primary Emissions of NO2 in Europe than Assumed in Policy Projections. *Nat. Geosci.* **2017**, *10* (12), 914–918. https://doi.org/10.1038/s41561-017-0009-0.

(4) Daellenbach, K. R.; Uzu, G.; Jiang, J.; Cassagnes, L.-E.; Leni, Z.; Vlachou, A.; Stefenelli, G.; Canonaco, F.; Weber, S.; Segers, A.; et al. Sources of Particulate-Matter Air Pollution and Its Oxidative Potential in Europe. *Nature* **2020**, *587* (7834), 414–419. https://doi.org/10.1038/s41586-020-2902-8.

(5) Reyna, J. L.; Chester, M. V; Ahn, S.; Fraser, A. M. Improving the Accuracy of Vehicle Emissions Profiles for Urban Transportation Greenhouse Gas and Air Pollution Inventories. *Environ. Sci. Technol.* **2015**, *49* (1), 369–376.

(6) Buckeridge, D. L.; Glazier, R.; Harvey, B. J.; Escobar, M.; Amrhein, C.; Frank, J. Effect of Motor Vehicle Emissions on Respiratory Health in an Urban Area. *Environ. Health Perspect.* **2002**, *110* (3), 293–300.

(7) Anenberg, S. C.; Miller, J.; Minjares, R.; Du, L.; Henze, D. K.; Lacey, F.; Malley, C. S.; Emberson, L.; Franco, V.; Klimont, Z. Impacts and Mitigation of Excess Diesel-Related NO x Emissions in 11 Major Vehicle Markets. *Nature* **2017**, *545* (7655), 467–471.

(8) Dedoussi, I. C.; Eastham, S. D.; Monier, E.; Barrett, S. R. H. Premature Mortality Related to United States Cross-State Air Pollution. *Nature* **2020**, *578* (7794), 261–265. https://doi.org/10.1038/s41586-020-1983-8.

(9) Bishop, G. A.; Stedman, D. H. A Decade of On-Road Emissions Measurements. *Environ. Sci. Technol.* **2008**, *42* (5), 1651–1656.

(10) Huang, Y.; Surawski, N. C.; Yam, Y.-S.; Lee, C. K. C.; Zhou, J. L.; Organ, B.; Chan, E. F. C. Re-Evaluating Effectiveness of Vehicle Emission Control Programmes Targeting High-Emitters. *Nat. Sustain.* **2020**, *3* (11), 904–907.

(11) Wu, Y.; Wang, R.; Zhou, Y.; Lin, B.; Fu, L.; He, K.; Hao, J. On-Road Vehicle Emission Control in Beijing: Past, Present, and Future. ACS Publications 2011.

(12) Tan, Y.; Henderick, P.; Yoon, S.; Herner, J.; Montes, T.; Boriboonsomsin, K.; Johnson, K.; Scora, G.; Sandez, D.; Durbin, T. D. On-Board Sensor-Based NO x Emissions from Heavy-Duty Diesel Vehicles. *Environ. Sci. Technol.* **2019**, *53* (9), 5504–5511.

(13) Zhu, M.; Dong, H.; Yu, F.; Liao, S.; Xie, Y.; Liu, J.; Sha, Q.; Zhong, Z.; Zeng, L.; Zheng, J. A New Portable Instrument for Online Measurements of Formaldehyde: From Ambient to Mobile Emission Sources. *Environ. Sci. Technol. Lett.* **2020**, *7* (5), 292–297.

(14) Davison, J.; Rose, R. A.; Farren, N. J.; Wagner, R. L.; Murrells, T. P.; Carslaw, D. C. Verification of a National Emission Inventory and Influence of On-Road Vehicle Manufacturer-Level Emissions. *Environ. Sci. Technol.* **2021**.

(15) Bishop, G. A.; Haugen, M. J. The Story of Ever Diminishing Vehicle Tailpipe Emissions as Observed in the Chicago, Illinois Area. *Environ. Sci. Technol.* **2018**, *52* (13), 7587–7593.

(16) Grange, S. K.; Farren, N. J.; Vaughan, A. R.; Davison, J.; Carslaw, D. C. Post-Dieselgate: Evidence of NOx Emission Reductions Using on-Road Remote Sensing. *Environ. Sci. Technol. Lett.* **2020**, *7* (6), 382–387.

(17) Chen, Y.; Sun, R.; Borken-Kleefeld, J. On-Road NOx and Smoke Emissions of Diesel Light Commercial Vehicles-Combining Remote Sensing Measurements from across Europe. *Environ. Sci. Technol.* **2020**, *54* (19), 11744–11752.

(18) Pujadas, M.; Domínguez-Sáez, A.; De la Fuente, J. Real-Driving Emissions of Circulating Spanish Car Fleet in 2015 Using RSD Technology. *Sci. Total Environ.* **2017**, *576*, 193–209.

(19) Ekström, M.; Sjödin, Å.; Andreasson, K. Evaluation of the COPERT III Emission Model with On-Road Optical Remote Sensing Measurements. *Atmos. Environ.* **2004**, *38* (38), 6631–6641.

(20) Chen, Y.; Borken-Kleefeld, J. Real-Driving Emissions from Cars and Light Commercial Vehicles–Results from 13 Years Remote Sensing at Zurich/CH. *Atmos. Environ.* **2014**, *88*, 157–164.

(21) Huang, Y.; Organ, B.; Zhou, J. L.; Surawski, N. C.; Hong, G.; Chan, E. F. C.; Yam, Y. S. Remote Sensing of On-Road Vehicle Emissions: Mechanism, Applications and a Case Study from Hong Kong. *Atmos. Environ.* **2018**, *182*, 58–74.

(22) Sun, K.; Tao, L.; Miller, D. J.; Pan, D.; Golston, L. M.; Zondlo, M. A.; Griffin, R. J.; Wallace, H. W.; Leong, Y. J.; Yang, M. M. Vehicle Emissions as an Important Urban Ammonia Source in the United States and China. *Environ. Sci. Technol.* **2017**, *51* (4), 2472–2481.

(23) Wu, Y.; Zhang, S.; Hao, J.; Liu, H.; Wu, X.; Hu, J.; Walsh, M. P.; Wallington, T. J.; Zhang, K. M.; Stevanovic, S. On-Road Vehicle Emissions and Their Control in China: A Review and Outlook. *Sci. Total Environ.* **2017**, *574*, 332–349.

(24) Huang, Y.; Organ, B.; Zhou, J. L.; Surawski, N. C.; Yam, Y.; Chan, E. F. C. Characterisation of Diesel Vehicle Emissions and Determination of Remote Sensing Cutpoints for Diesel High-Emitters. *Environ. Pollut.* **2019**, *252*, 31–38.

(25) Grange, S. K.; Farren, N. J.; Vaughan, A. R.; Rose, R. A.; Carslaw, D. C. Strong Temperature Dependence for Light-Duty Diesel Vehicle NO x Emissions. *Environ. Sci. Technol.* **2019**, *53* (11), 6587–6596.

(26) Carslaw, D. C.; Farren, N. J.; Vaughan, A. R.; Drysdale, W. S.; Young, S.; Lee, J. D. The Diminishing Importance of Nitrogen Dioxide Emissions from Road Vehicle Exhaust. *Atmos. Environ. X* **2019**, *1*, 100002.

(27) Huang, Y.; Ng, E. C. Y.; Surawski, N. C.; Yam, Y.-S.; Mok, W.-C.; Liu, C.-H.; Zhou, J. L.; Organ, B.; Chan, E. F. C. Large Eddy Simulation of Vehicle Emissions Dispersion: Implications for on-Road Remote Sensing Measurements. *Environ. Pollut.* **2020**, *259*, 113974.

(28) Jimenez, J. L.; McClintock, P.; McRae, G. J.; Nelson, D. D.; Zahniser, M. S. Vehicle Specific Power: A Useful Parameter for Remote Sensing and Emission Studies. In *Ninth CRC On-Road Vehicle Emissions Workshop, San Diego, CA*; 1999.

(29) Smit, R.; Kingston, P.; Neale, D. W.; Brown, M. K.; Verran, B.; Nolan, T. Monitoring On-Road Air Quality and Measuring Vehicle Emissions with Remote Sensing in an Urban Area. *Atmos. Environ.* **2019**, *218*, 116978.

(30) Kang, Y.; Li, Z.; Lv, W.; Xu, Z.; Zheng, W. X.; Chang, J. High-Emitting Vehicle Identification by on-Road Emission Remote Sensing with Scarce Positive Labels. *Atmos. Environ.* **2021**, *244*, 117877.

(31) Anthony, M.; Bartlett, P. L. *Neural Network Learning: Theoretical Foundations*; cambridge university press, 2009.

(32) Pal, M. Random Forest Classifier for Remote Sensing Classification. *Int. J. Remote Sens.* **2005**, *26* (1), 217–222.

(33) Chen, T.; He, T.; Benesty, M.; Khotilovich, V.; Tang, Y.; Cho, H. Xgboost: Extreme Gradient Boosting. *R Packag. version 0.4-2* **2015**, *1* (4).

(34) Requia, W. J.; Di, Q.; Silvern, R.; Kelly, J. T.; Koutrakis, P.; Mickley, L. J.; Sulprizio, M. P.; Amini, H.; Shi, L.; Schwartz, J. An Ensemble Learning Approach for Estimating High Spatiotemporal Resolution of Ground-Level Ozone in the Contiguous United States. *Environ. Sci. Technol.* **2020**, *54* (18), 11037–11047.

(35) Shtein, A.; Kloog, I.; Schwartz, J.; Silibello, C.; Michelozzi, P.; Gariazzo, C.; Viegi, G.; Forastiere, F.; Karnieli, A.; Just, A. C. Estimating Daily PM2.5 and PM10 over Italy Using an Ensemble Model. *Environ. Sci. Technol.* **2019**, *54* (1), 120–128.

(36) Di, Q.; Amini, H.; Shi, L.; Kloog, I.; Silvern, R.; Kelly, J.; Sabath, M. B.; Choirat, C.; Koutrakis, P.; Lyapustin, A. Assessing NO2 Concentration and Model Uncertainty with High Spatiotemporal Resolution across the Contiguous United States Using Ensemble Model Averaging. *Environ. Sci. Technol.* **2019**, *54* (3), 1372–1384.

(37) Wolpert, D. H. Stacked Generalization. *Neural networks* **1992**, *5* (2), 241–259.

(38) Chang, P.-C. A Novel Model by Evolving Partially Connected Neural Network for Stock Price Trend Forecasting. *Expert Syst. Appl.* **2012**, *39* (1), 611–620.

(39) Srivastava, N.; Hinton, G.; Krizhevsky, A.; Sutskever, I.; Salakhutdinov, R. Dropout: A Simple Way to Prevent Neural Networks from Overfitting. *J. Mach. Learn. Res.* **2014**, *15* (1), 1929–1958.

(40) Probst, P.; Wright, M. N.; Boulesteix, A. Hyperparameters and Tuning Strategies for Random Forest. *Wiley Interdiscip. Rev. Data Min. Knowl. Discov.* **2019**, *9* (3), e1301.

(41) Pedregosa, F.; Varoquaux, G.; Gramfort, A.; Michel, V.; Thirion, B.; Grisel, O.; Blondel, M.; Prettenhofer, P.; Weiss, R.; Dubourg, V. Scikit-Learn: Machine Learning in Python. *J. Mach. Learn. Res.* **2011**, *12*, 2825–2830.

(42) Zhang, K.; Zhong, S.; Zhang, H. Predicting Aqueous Adsorption of Organic Compounds onto Biochars, Carbon Nanotubes, Granular Activated Carbons, and Resins with Machine Learning. *Environ. Sci. Technol.* **2020**, *54* (11), 7008–7018.

(43) Sundararajan, M.; Najmi, A. The Many Shapley Values for Model Explanation. In *International Conference on Machine Learning*; PMLR, 2020; pp 9269–9278.

(44) Jiang, P.; Chen, X.; Li, Q.; Mo, H.; Li, L. High-Resolution Emission Inventory of Gaseous and Particulate Pollutants in Shandong Province, Eastern China. *J. Clean. Prod.* **2020**, *259*, 120806. https://doi.org/https://doi.org/10.1016/j.jclepro.2020.120806.

(45) Sun, S.; Jin, J.; Xia, M.; Liu, Y.; Gao, M.; Zou, C.; Wang, T.; Lin, Y.; Wu, L.; Mao, H.; et al. Vehicle Emissions in a Middle-Sized City of China: Current Status and Future Trends. *Environ. Int.* **2020**, *137*, 105514. https://doi.org/https://doi.org/10.1016/j.envint.2020.105514.

(46) Chen, Y.; Zhang, Y.; Borken-Kleefeld, J. When Is Enough? Minimum Sample Sizes for on-Road Measurements of Car Emissions. *Environ. Sci. Technol.* **2019**, *53* (22), 13284–13292.

(47) Bishop, G. A.; DeFries, T. H.; Sidebottom, J. A.; Kemper, J. M. Vehicle Exhaust Remote Sensing Device Method to Screen Vehicles for Evaporative Running Loss Emissions. *Environ. Sci. Technol.* **2020**, *54* (22), 14627–14634.

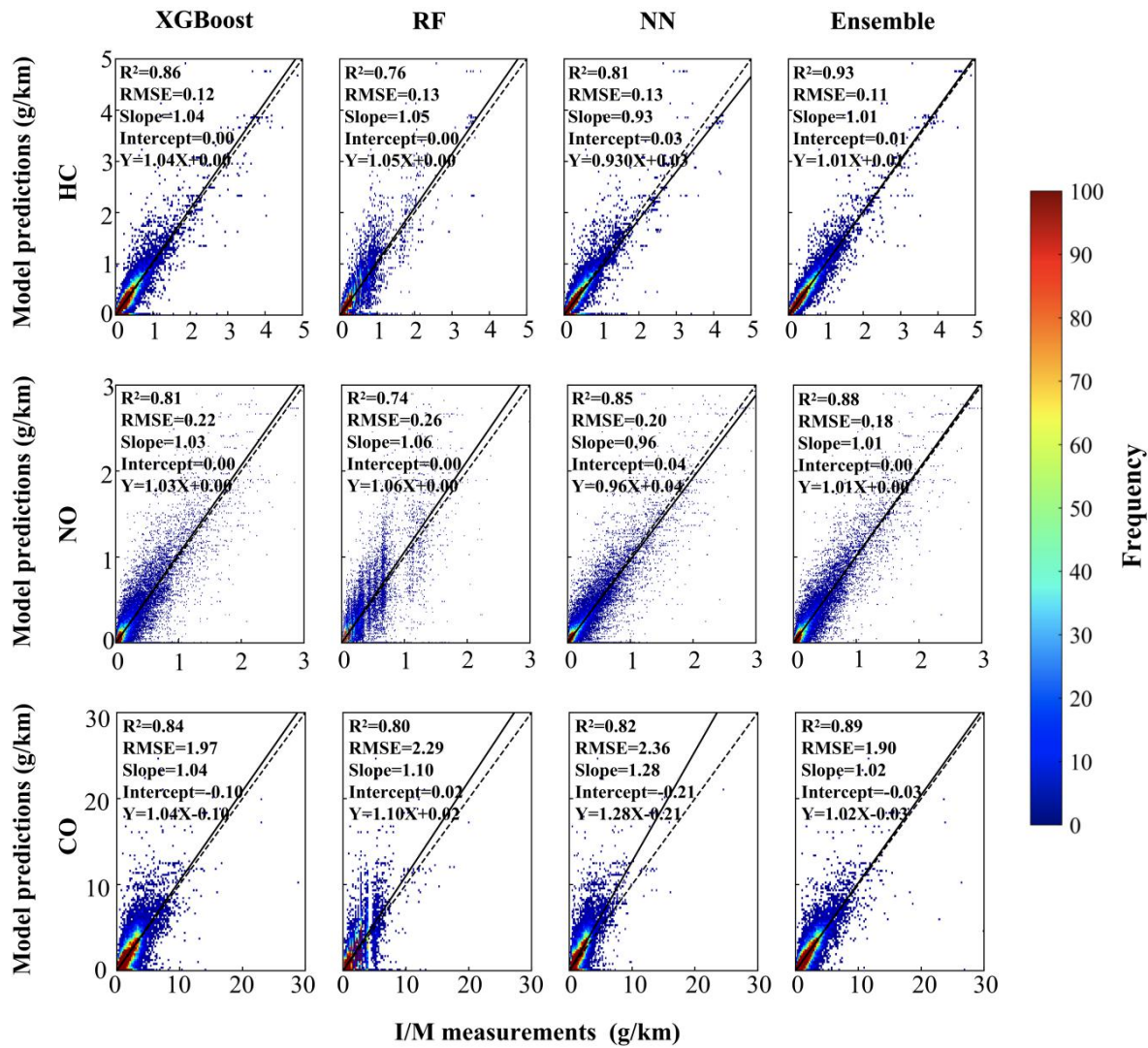

**Figure 1. Density scatterplots of the cross-validation of the predicted vehicle-specific emissions (including CO, HC, and NO) via FNN, RF, XGBoost, and the ensemble model for the validation dataset.** The statistical metrics (including $R^2$, RMSE, Slop, and Intercept) and the linear regression relationship are presented in each panel. The color represents the frequency for the number of data points. The red dashed lines denote the best-fit lines through the data points.

Submitted to Environmental Science & Technology

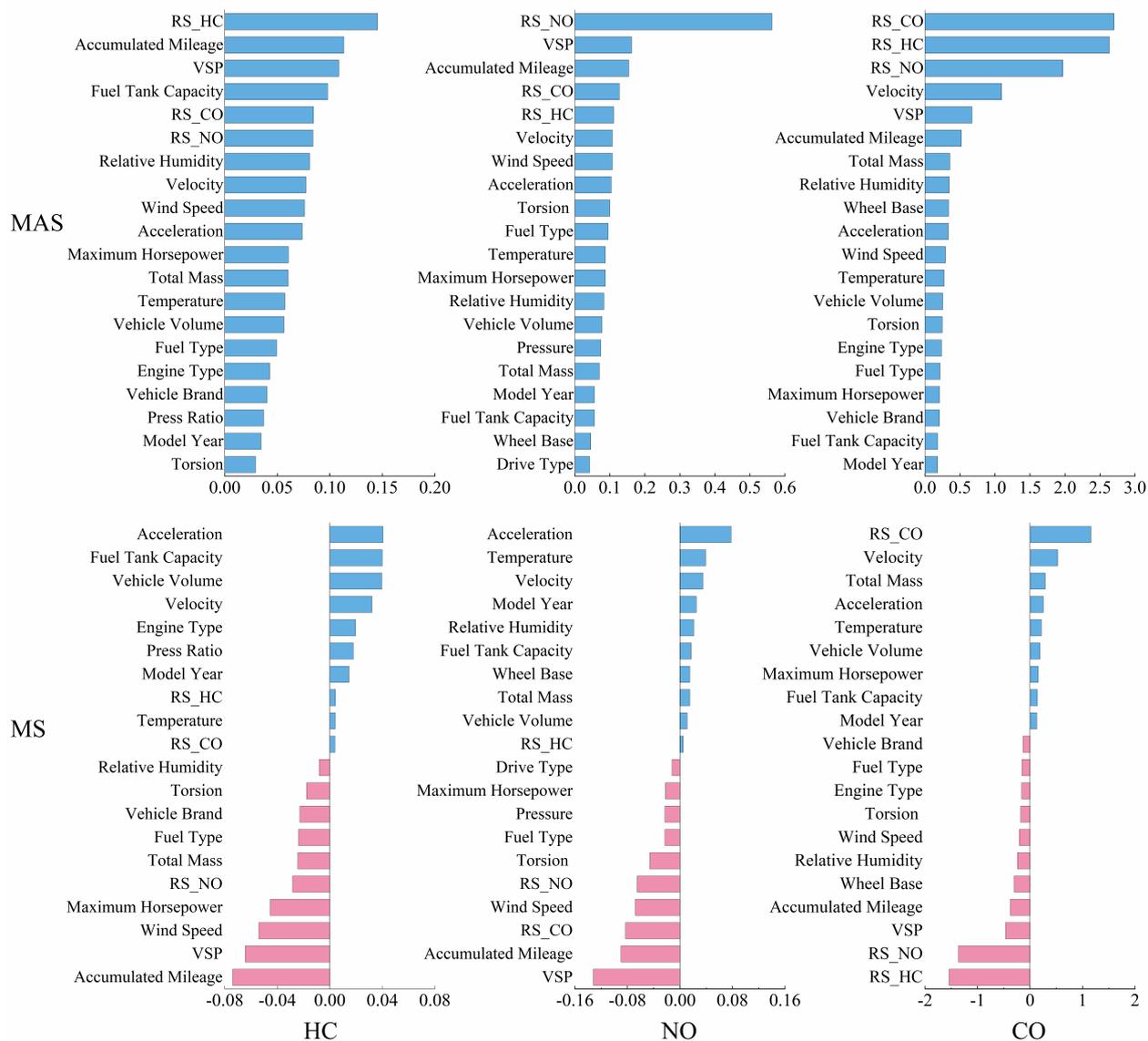

**Figure 2. MS and MAS values for input variables for the predicted vehicle-specific emissions (i.e., CO, HC, and NO) via the ensemble model.** All input variables are recorded in Table S2. The input variables with the top 20 impacts on the predicted emissions are shown here. Therein fuel tank capacity, vehicle volume, engine type, press ratio, model year, torsion, vehicle brand, fuel type, maximum horsepower, total mass, and accumulated mileage are obtained from the I/M databases, while RS_HC, RS_CO, RS_NO, velocity, acceleration, VSP, temperature, relative humidity and wind speed come from the ORRS measurements.

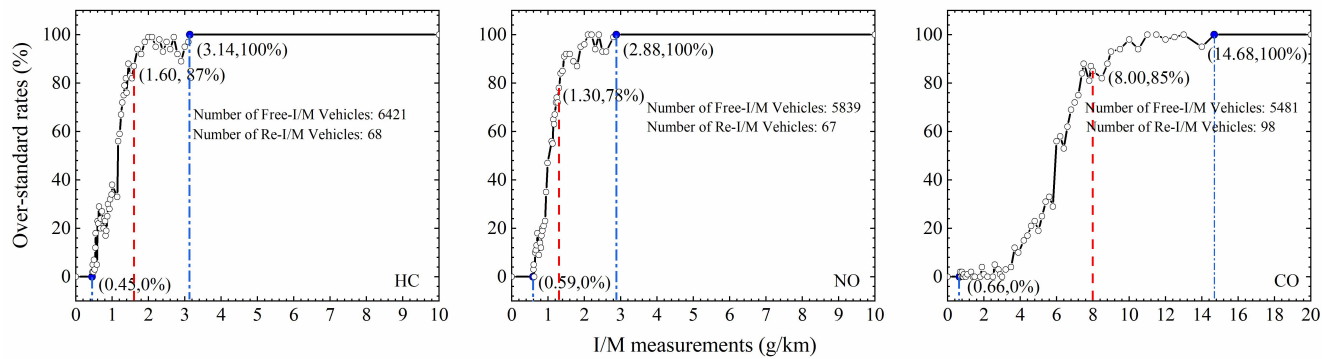

**Figure 3. The over-standard rates of the predicted vehicle-specific emissions (i.e., CO, HC, and NO) across the I/M-based emission magnitude.** For a certain pollutant, there are three intervals divided by two thresholds (blue dash lines). The former interval corresponds to the Free-I/M vehicles, while the latter one the Re-I/M vehicles. The red dash lines refer to the official emission standards. The numbers of the Free-I/M and Re-I/M vehicles are also presented.

Submitted to Environmental Science & Technology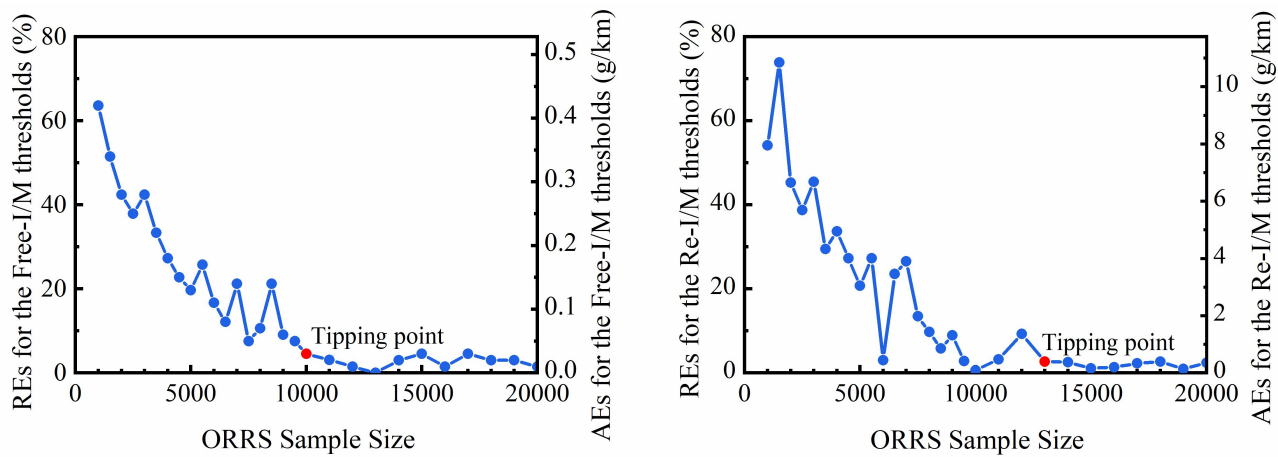

**Figure 4. ORRS sample sizes vs the REs and AEs of the predicted Free-I/M and Re-I/M thresholds.** The **RE**s and **AE**s of the predicted NO thresholds show as a function of the ORRS sample sizes. A larger sample size leads to higher confidence in predicting the thresholds. More importantly, we find the tipping points (Red points) in this function, below which further increases in the sample size lead to remarkable decreases in the **RE**s and **AE**s. Specifically, a sample size of approximately 10000 records is needed for the predicted Free-I/M thresholds, while a significantly large size of 14000 records for the predicted Re-I/M thresholds.

22